\documentclass{article}
\usepackage{jfrExamplee}
\usepackage{graphicx}
\usepackage{apalike}
\usepackage{setspace}
\usepackage{amsmath}
\usepackage{xcolor}

\title{Discovery of Skill Switching Criteria for Learning Agile Quadruped Locomotion}

\author{
Wanming Yu
\\
Oxford Robotics Institute\\
University of Oxford\\
\texttt{wanming@robots.ox.ac.uk} \\
\And
Fernando Acero \\
Department of Computer Science \\
University College London\\
\texttt{fernando.acero@ucl.ac.uk} \\
\AND
Vassil Atanassov \\
Oxford Robotics Institute\\
University of Oxford\\
\texttt{vassilatanassov@robots.ox.ac.uk} \\
\And
Chuanyu Yang \\
Amigaga Technology Co. Ltd \\
\texttt{ycy0char@gmail.com} \\
\And
Ioannis Havoutis \\
Oxford Robotics Institute\\
University of Oxford\\
\texttt{ioannis@robots.ox.ac.uk} \\
\And
Dimitrios Kanoulas \\
Department of Computer Science \\
University College London\\
\texttt{d.kanoulas@ucl.ac.uk} \\
\And
Zhibin Li \\
Department of Computer Science \\
University College London\\
\texttt{alex.li@ucl.ac.uk} \\
}

\begin{document}

\maketitle

\begin{abstract}
This paper develops a hierarchical learning and optimization framework that can learn and achieve well-coordinated multi-skill locomotion. The learned multi-skill policy can switch between skills automatically and naturally in tracking arbitrarily positioned goals and recover from failures promptly.
The proposed framework is composed of a deep reinforcement learning process and an optimization process. First, the contact pattern is incorporated into the reward terms for learning different types of gaits as separate policies without the need for any other references. 
Then, a higher level policy is learned to generate weights for individual policies to compose multi-skill locomotion in a goal-tracking task setting. Skills are automatically and naturally switched according to the distance to the goal. The proper distances for skill switching are incorporated in reward calculation for learning the high level policy and updated by an outer optimization loop as learning progresses.
We first demonstrated successful multi-skill locomotion in comprehensive tasks on a simulated Unitree A1 quadruped robot. We also deployed the learned policy in the real world showcasing trotting, bounding, galloping, and their natural transitions as the goal position changes. Moreover, the learned policy can react to unexpected failures at any time, perform prompt recovery, and resume locomotion successfully.
Compared to discrete switch between single skills which failed to transition to galloping in the real world, our proposed approach achieves all the learned agile skills, with smoother and more continuous skill transitions. 

\textbf{Keywords:} Multi-skill locomotion; Deep reinforcement learning; Hierarchical learning and optimization; Gait transitions; Skill switching; Legged locomotion; Robot learning 
\end{abstract}

\section{Introduction}
\begin{figure}[h]
  \centering
  \includegraphics[height=2.0in]{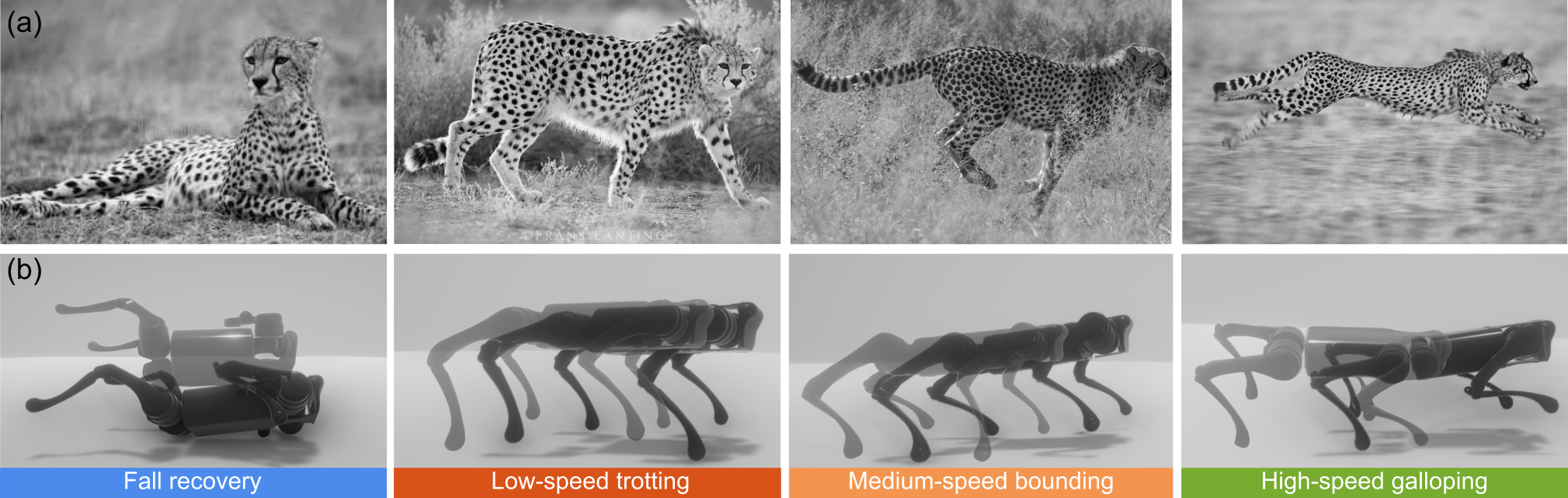}
  \caption{Coordination and gait transitions in quadrupedal animal and robot on demand of increasing speed. (a) Cheetah's changing gaits at increasing speed. (b) \textit{A1} quadruped robot's fall recovery, trotting, bounding and galloping skills using our multi-skill policy.}
  \label{fig:highlight}
\end{figure}

Animals have evolved highly efficient movement strategies. Mimicking these can improve legged locomotion in terms of agility, stability, and adaptivity (Fig. \ref{fig:highlight}). In particular, animals learn to switch between motor skills swiftly according to tasks and surroundings. For instance, horses switch to different gait patterns as speed changes \cite{hoyt1981gait}. However, reproducing multiple gaits and their dynamically feasible transitions on legged robots remains challenging in the robot learning and control community. 
Besides, the ability to recover from various failures, which is of vital interest for successful and resilient real-world deployment, is not yet well studied in multi-skill locomotion.
A multi-skill framework has the ability to recover from failures during locomotion \cite{yang2020multi}, however, it does not show more dynamic gaits other than trotting. Although fall recovery has been studied in several previous works \cite{cite:hwangbo2019learningAgile,castano2019design}, it is learned as a single skill but cannot be combined with other skills.

\textbf{Reproducing gait patterns.} Behavior cloning or imitation learning approaches have been applied to reproduce various gaits from reference motions. Reference motions can be captured from animal locomotion which can be limited in variety or generated from model predictive controllers \cite{reske2021imitation} which require domain knowledge and computing efficiency. In general, such approaches can be difficult to scale beyond the dataset, lacking robustness and diversity of the learned behaviors.
There have been attempts to model or learn parameterized control policy to achieve different styles of walking. A phase-guided controller is used to learn gait transitions between walking, trotting, pacing and bounding on a quadrupedal robot \cite{shao2021learning}.
A single policy is learned to control various gaits with variable footswing, posture and speed \cite{margolis23a}. However, such approaches need to handcraft control-specific behavior parameters, which need domain knowledge and are not intuitive to switch gaits.

\textbf{Generative models in multi-skill locomotion.} Recent advances in generative models have achieved low-level control of locomotion gaits on quadrupedal robots. 
Variational autoencoders have been applied to learn a disentangled and 2D latent representation across locomotion gaits with respect to footswing heights and lengths \cite{mitchell2024gaitor}. Given desired gait type and swing characteristics, it achieved the low-level control of trotting, crawling and pacing on quadrupedal robots.
Besides, diffusion models have shown the capabilities of achieving multi-skill locomotion control by a single policy, such as trotting, hopping, pacing, walking and running \cite{huang2024diffuseloco}, and walking, crawling and their transitions \cite{o2024offline}. However, these generative models require expert demonstrations, and the performance is dependent on the quality of the dataset. Moreover, the gait type was conditioned on a certain input to the control policy. In contrast, the gait type was autonomously discovered by our proposed hierarchical framework covering both high-level and low-level multi-skill locomotion control.

\textbf{Foundation models in legged locomotion.} Applying foundation models in robot learning applications becomes an attractive approach to achieve generalized robot tasks and behaviours. Pre-trained vision-language models (VLMs) usually focus on high-level reasoning and planning to select from a set of existing low-level skills \cite{chen2024commonsense}. 
However, certain low-level locomotion skills can be difficult to obtain in practice.
Several attempts have also been made to apply pre-trained Large Language Models (LLMs) to achieve multiple gaits via low-level interfaces, such as foot contact patterns \cite{tang2023saytap}. 
In general, these foundation models in robotic applications require either careful prompt engineering or a huge amount of robotic data to fine-tune. In practice, the robotic data can be difficult to obtain in certain cases and the fine-tuning of large-scale models may require heavy computing resources.

\textbf{Bio-inspired multi-gait locomotion.} Unlike the above approaches requiring reference motions, there has been robotics research applying deep reinforcement learning approaches to acquire animal-like gait transitions according to various criteria inspired by biological principles, where reference motions are not necessary.
By minimizing energy consumption, the robot can achieve gait transitions between walking, trotting and fly-trotting at different speed ranges using a single policy \cite{liang2024adaptive} or a hierarchical structure \cite{yang2022fast}, and gait transitions from walking to trotting to bouncing \cite{fu2021minimizing}.
Another bio-inspired research modulates gait transitions according to Froude numbers \cite{humphreys2023bio}.
A more recent work learned gait transitions from walking to trotting on a flat ground and trotting to pronking when crossing gaps according to viability \cite{shafiee2024viability}. 
However, galloping cannot emerge or be incorporated at high speed on mechanical robots in the above frameworks.
Moreover, a series of work utilizes Central Pattern Generators (CPGs) to produce different gaits by deep reinforcement learning on quadrupedal robots.
The most recent work took a coupling-driven approach to learn a policy by deep reinforcement learning to modulate the parameters of Central Pattern Generators (CPGs), where nine gaits and transitions including galloping have been produced according to Cost of Transport (CoT) \cite{bellegarda2024allgaits}.
In contrast with bio-inspired multi-gait locomotion, our proposed framework can produce galloping gait. Moreover, our approach is not constrained by biological principles as we can define customized cost terms for optimizing gait switch timing. Besides biology-inspired criteria, we can also include other cost terms, such as task-related costs.

\textbf{State-of-the-art quadrupedal locomotion.} A parallel line of research demonstrated impressive dynamic parkour skills on legged robots \cite{caluwaerts2023barkour,cheng2023extreme,zhuang2023robot,he2024agile}. However, these works usually focused on navigating the robot along a series of challenging terrains and obstacles. In most cases, a simple goal-reaching task is considered in these navigation tasks, while gait patterns are not taken into account. In contrast, this paper focuses on multi-skill navigation and control tasks, i.e., reaching arbitrary goals with various gait patterns and their transitions.

\textbf{Contributions.} To summarize, the advantages of our proposed approach over the existing literature include: (1) we do not require reference trajectories or expert demonstrations, and can learn multi-skill locomotion purely from scratch; (2) animal-like galloping gait can be activated at high-speed locomotion; (3) autonomous fall recovery is incorporated in the multi-skill policy, enabling high robustness and requiring less human intervention; (4) flexible gait switch criteria are automatically discovered for mechanical robots.
To the best of our knowledge, our work is the first multi-skill learning and optimization framework that is compatible to incorporate and synthesize multiple highly dynamic locomotion gaits (especially galloping), and produce natural, dynamically feasible transitions by automatically discovered gait switch criteria. Our work demonstrates four skills on a quadruped robot in the real world including prompt fall recovery at any stage during multi-skill locomotion. To summarize, the contributions of this paper include:
\begin{itemize}
    \item Incorporate highly dynamic locomotion skills of bounding and galloping besides trotting into learning one coherent multi-skill policy, without the need for reference trajectories.
    \item Demonstrate successful trotting, bounding and galloping and their dynamically feasible and continuous transitions with one synthesized multi-skill policy on a real quadruped robot.
    \item Successful failure recovery at any stage of different gaits.
    \item Automatic discovery of gait switch criteria as motor learning progresses.
\end{itemize}

\begin{figure}[h]
  \centering
  \includegraphics[height=1.8in]{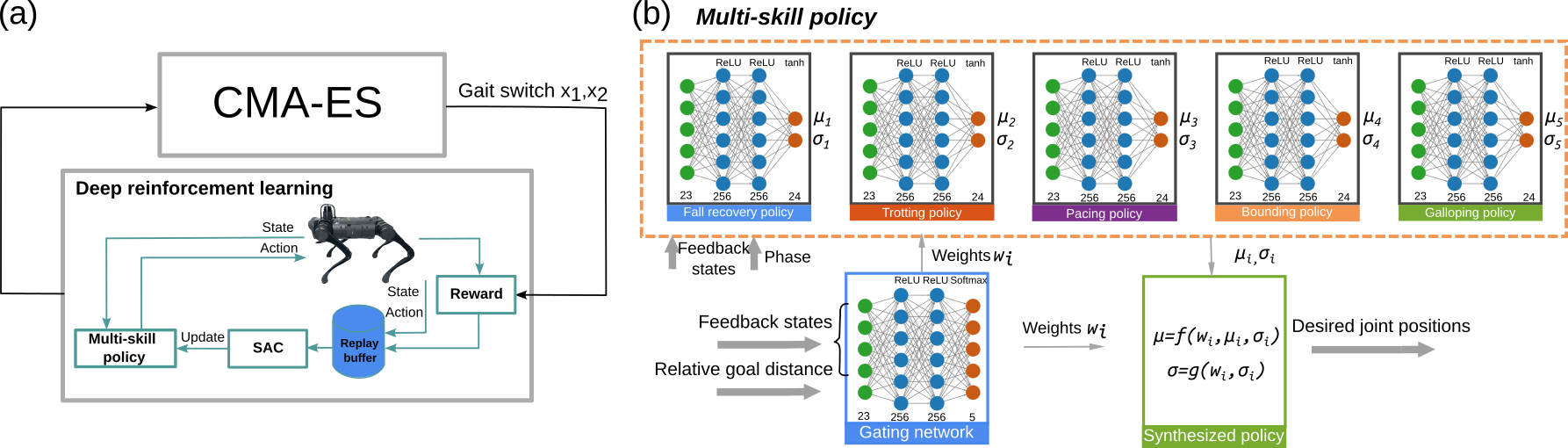}
  \caption{Proposed multi-skill learning and optimization framework. (a) Optimizing gait switch criteria in the outer loop of deep reinforcement learning. (b) Neural network architecture of a multi-skill policy. Bold arrows are the input or output outside the policy, while normal arrows are the internal input or output.}
  \label{fig:flowchart}
\end{figure}

Our hierarchical multi-skill learning and optimization framework is shown in Fig. \ref{fig:flowchart}, which includes: (1) a set of pre-trained reusable single-skill neural network policies, each representing a single locomotion skill; (2) a task-level neural network which generates weights for each skill to produce our multiplicative composite policy; (3) the composite multi-skill policy; and (4) the outer optimization loop for the discovery of gait switch criteria represented by the relative goal distances in the horizontal plane that activate switch from trotting and bounding, and switch from bounding to galloping, respectively.

In the following sections, we will first review the details of our hierarchical learning and optimization framework in Section \ref{method1} and \ref{method2}. Then we demonstrate and analyze the learned multi-skill locomotion policy on a real quadruped robot in Section \ref{results}. Finally, we conclude our work in Section \ref{conclusion}.

\section{Hierarchical multi-skill learning framework}
\label{method1}

\subsection{Learning individual skills}
The robot learns five individual skills separately using a systematic deep reinforcement learning framework, including fall recovery, trotting, pacing, bounding and galloping. Each locomotion skill is a feedback control policy represented by a neural network, which is learned by the Soft Actor-Critic (SAC) algorithm \cite{haarnoja2018soft}. Details of the key components for our deep reinforcement learning framework are given below. 

\subsubsection{State observation and action space}
Following the key feedback states in learning locomotion skills \cite{yu2023identifying}, the state input to the Actor neural network includes: (1) normalized gravity vector in the robot local frame which reflects the body orientation of the robot, (2) base angular velocity, (3) base linear velocity in the robot heading frame, and (4) joint positions.  
For learning periodic locomotion skills, we also included a two-dimensional phase vector $(sin2\pi\phi,cos2\pi\phi)$ to represent continuous temporal information that encodes phase $\phi$ from 0\% to 100\% of a periodic motion.
The actions are the desired joint positions for twelve joints, including hip roll, hip pitch and knee joints of four legs.

\subsubsection{Reward design}
Trotting and bounding were learned with a fixed desired velocity, while galloping was learned by maximizing velocity. The reward function for learning individual policies are composed of continuous and discrete reward terms. For continuous reward terms, we use a radial basis function (RBF) to formulate as follows:
\begin{equation}
\varphi(x, \widehat{x}, \alpha)=\exp \left(\alpha(\hat{x}-x)^{2}\right)   
\end{equation}
where $x$ is the continuous physical quantity, $\hat{x}$ is the corresponding reference, and $\alpha$ is the shape parameter which controls the width of RBF.
Formulation and weight of each reward term are given in Table \ref{reward} and \ref{weight}, respectively.

\begin{table}[h]
\caption{Reward terms for learning quadruped locomotion skills.} \label{reward}
\begin{center}
\begin{tabular}{|l|l|}
  \hline
    Physical quantities& Reward term \\
  \hline\hline
    Base orientation& $w_{\phi}\times\varphi(\phi, [0,0,-1], -2.35)$ \\ \hline
    Base height& $w_h\times\varphi(h, \hat{h}, -51.16)$ \\ \hline
    Base linear velocity& $w_v\times\begin{cases}

    {{v_x}_{base}^{world}}^2, &\text{gallop}\\
    \varphi(v_{base}^{world}, \hat{v}_{base}^{world}, -18.42), &\text{else}
    \end{cases}
    $ \\ \hline
    Joint torque& $w_{\tau}\times\varphi(\tau, 0, -0.004)$          \\ \hline
    Joint velocity& $w_{\dot{q}}\times\varphi(\dot{q}, 0, -0.032)$ \\ \hline
    Body ground contact& $
    w_{bg}\times
    \begin{cases}
    0, &\text{base in contact with ground}\\
    1, &\text{base not in contact with ground}
    \end{cases}
    $ \\ \hline
    Foot ground contact& $
    w_{fg}\times
    \begin{cases}
    0, &\text{no foot in contact with ground}\\
    1, &\text{foot in contact with ground}
    \end{cases}
    $ \\ \hline
    Symmetric foot placement &
    $w_{pf}\times
    \begin{cases}
    \varphi(p_{foot}^{base}, \hat{p}_{foot}^{base}, -51.16), &\text{recovery}\\
    \varphi(1/4 \sum_{n=1}^{4}(p_{foot, n}^{world}), p_{base}^{world}, -51.16), &\text{gaits}
    \end{cases}
    $ \\ \hline
    Swing and stance &
    $w_{hf}\times\varphi(h_{foot}^{world}v_{foot}^{world},\hat{h}_{foot}^{world}v_{foot}^{world}, -460.50)$ \\ \hline
    Yaw velocity& $w_{\dot{\psi}}\times\varphi(\dot{\psi}, 0, -7.47)$          \\ \hline
        Reference foot contact& $
    w_f\times\begin{cases}
    0, &\text{not match desired foot contact}\\
    1, &\text{match desired foot contact}
    \end{cases}
    $ \\ 
  \hline
\end{tabular}
\end{center}
\end{table}

\begin{table}[h]
\caption{Reward term weights for learning single locomotion skills.} \label{weight}
\begin{center}
\begin{tabular}{|c|c|c|c|c|c|c|c|c|c|c|c|}
  \hline
    Task & $w_{\phi}$ & $w_h$ & $w_v$ & $w_{\tau}$& $w_{\dot{q}}$& $w_{bg}$& $w_{fg}$& $w_{pf}$&$w_{hf}$&$w_{\dot{\psi}}$&$w_f$ \\
  \hline\hline
   Fall recovery  & 0.189&0.189&0.114&0.076&0.076&0.083&0.083&0.189&0.000&0.000&0.000\\ \hline
   Gaits & 0.068&0.068&0.170&0.017&0.017&0.048&0.000&0.034&0.034&0.068&0.476\\
    \hline
\end{tabular}
\end{center}
\end{table}

\subsubsection{Reference foot contact reward}
\begin{figure}[h]
  \centering
  \includegraphics[height=2.0in]{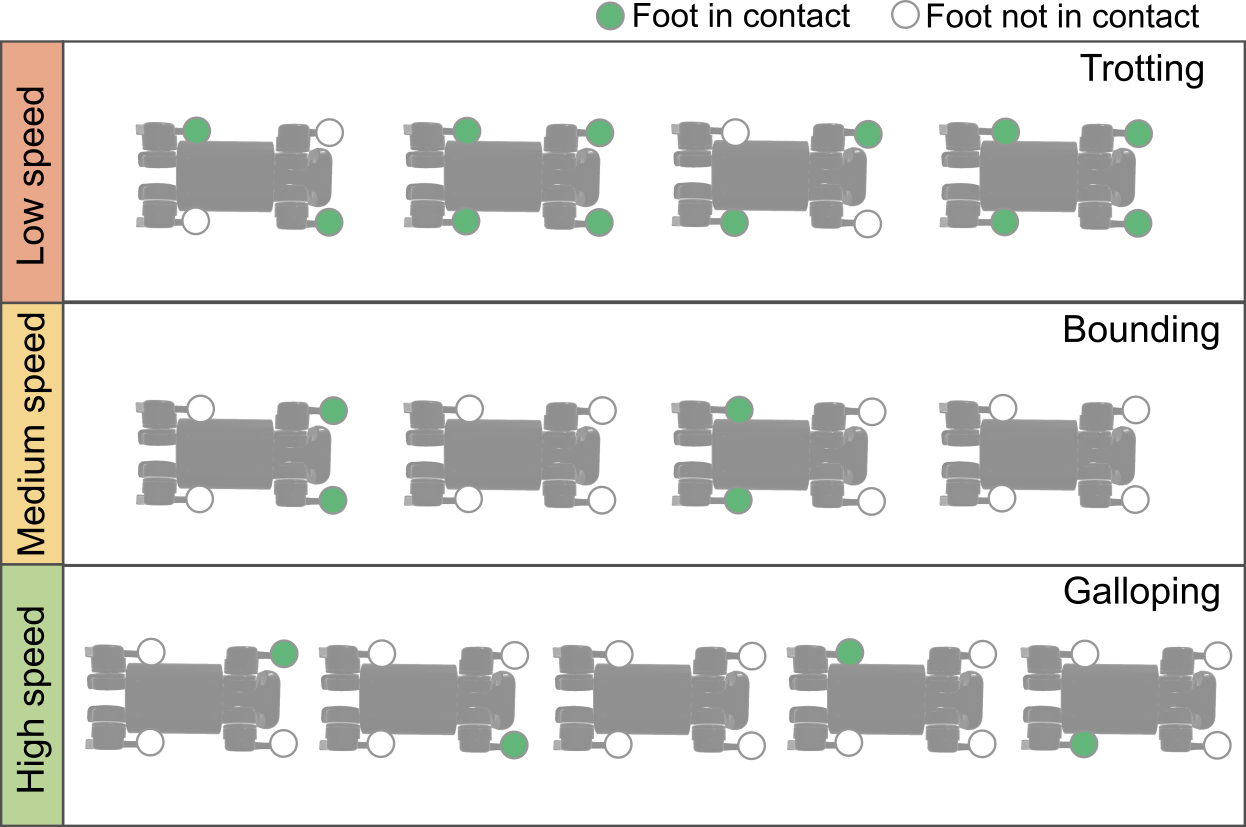}
  \caption{Foot contact patterns.}
  \label{fig:contact}
\end{figure}

The last reward term in Table \ref{reward}, i.e., reference foot contact reward, is the key to learning different gait types without reference. 
The desired foot contact pattern for each gait type is inspired by quadrupedal animals \cite{owaki2017quadruped}, as shown in Fig. \ref{fig:contact}. In this paper, we assume trotting, bounding and galloping gait as speed increases as a proof-of-concept. It should be noted that the order of the gait types is not fixed. Here we determine the gait type at different stages according to its characteristics. 

\subsection{Learning multi-skill locomotion}
The task objective of the robot is to track an arbitrary goal in the horizontal plane with natural gait transitions from trotting to bounding to galloping. The goal is represented by $(d_g,\theta_g)$, and the goal position in the horizontal plane is ($d_g cos\theta_g$,$d_g sin\theta_g$). At each episode, we set a fixed goal, where $d_g\sim(0m,15m)$  and $\theta_g\sim(-180^\circ,180^\circ)$.
The multi-skill locomotion policy is synthesized on-the-fly by multiplicative composition \cite{peng2019mcp} of the low-level pre-trained individual policies according to the output from the high-level gating network. We will introduce more details about each component in the following sections.
\subsubsection{Gating network}
Our high-level gating network is a fully connected neural network with two hidden layers. Each hidden layer has 256 neurons and uses a ReLU activation function. The gating network receives the following as input: gravity vector, base angular velocity, base linear velocity, joint positions, plus normalized distance between the robot and the goal in the horizontal plane. The gating network outputs the weights for each locomotion skill which are added up to one.

\subsubsection{Composite multi-skill policy}
During the training of multi-skill locomotion, the parameters of the expert networks are transferred from the pre-trained single-skill policies and are fixed all the time. That is, only gating network parameters are updated by backpropagation of the gradient obtained from the designed reward.
The weights for each locomotion skill generated by the gating network are then applied to synthesize the multi-skill Gaussian policy $\pi(a \mid s, g)$ by the multiplicative composition of $n$ pre-trained single skills:
\begin{equation}
    \pi(a \mid s, g)=\frac{1}{Z(s, g)} \prod_{i=1}^{n} \pi_{i}(a \mid s)^{w_{i}(s, g)}, \quad w_{i}(s, g) \geq 0
\end{equation}
    
where $\pi_{i}(a \mid s)$ is the $i$-th single-skill neural network policy, $w_i(s,g)$ is the weight for the corresponding skill to influence the composite policy, and $Z(s, g)$ is the normalization factor.
The synthesized policy is a multiplicative composition of $n$ Gaussian policies, i.e., $n$ single skills. As discussed in \cite{peng2019mcp}, the multiplicative composition of Gaussian primitives results in another Gaussian policy, i.e., the composite policy. Due to the use of Gaussian primitives, the mean $\mu^1, \mu^2,...,\mu^{12}$ and standard deviation $\sigma^1, \sigma^2,...,\sigma^{12}$ of the synthesized Gaussian policy are obtained as follows:
\begin{equation}
\mu^{j}(s, g)=\frac{1}{\sum_{l=1}^{n} \frac{w_{l}(s, g)}{\sigma_{l}^{j}(s, g)}} \sum_{i=1}^{n} \frac{w_{i}(s, g)}{\sigma_{i}^{j}(s, g)} \mu_{i}^{j}(s, g)
\end{equation}
\begin{equation}
    \sigma^{j}(s, g)=\left(\sum_{i=1}^{n} \frac{w_{i}(s, g)}{\sigma_{i}^{j}(s, g)}\right)^{-1}
\end{equation}

It should be noted that we adopted a multiplicative model instead of an additive model in a hierarchical learning framework, such as Mixture of Experts (MoE) \cite{jacobs1991adaptive}, to avoid conflicting behaviors or blending artifacts caused by the sum of primitives, as reported in \cite{peng2019mcp}. 

\subsubsection{Control framework}
The pre-trained single-skill policies, the gating network, and the composite multi-skill policy run at 25 Hz. These policies together generate desired joint positions which are tracked by joint-level PD controllers at 1000 Hz. The PD controllers receive desired joint positions $\hat{q}$, measured joint positions $q$ and joint velocities $\dot{q}$ as input and output the joint torque commands $\tau = K_p(\hat{q}-q)+K_d(0-\dot{q})$.

\subsubsection{Reward design}
For multi-skill learning, we design three groups of reward terms. 
The most important group of reward terms is related to target-following $r_g$, another group is reference foot contact reward $r_f$, and the last group includes the rest of the reward terms in learning single locomotion skills $r_e$.
We set the overall reward $r$ as the weighted sum of these terms as follows:
\begin{equation}
    r = 0.6 r_g + 0.2 r_f + 0.2 r_e.
\end{equation}

\textbf{Goal-tracking reward:}
Our goal-tracking reward consists of three terms.
First, the relative position reward $r_{p_g}$, which encourages minimization of the relative distance between the robot and the goal in the horizontal plane $d\geq 0$:
\begin{equation}
    r_{p_g} = \varphi(p_{goal}^{world}, p_{base}^{world}, -0.74);
\end{equation}
Second, the robot velocity reward $r_{v_g}$,where robot velocity towards the goal in the horizontal plane is expected to be as fast as possible:
\begin{equation}
    r_{v_g} = v^2
\end{equation}
 
Third, the robot heading reward $r_{\phi_g}$, which encourages alignment of the robot heading towards the goal:
\begin{equation}
    r_{\phi_g} = \varphi(u_{goal,base}^{base}, [1,0,0], -2.35)
\end{equation}
where $u_{goal,base}^{base}$ is the unit vector pointing from the robot base to the goal in base frame.
Our full target-following reward is $r_g=r_{h_z} r_{\phi}(8 r_{p_g} + 4 r_{v_g} + 4 r_{\phi_g}) $, which increases the target-following reward weight when the robot is closer to the nominal standing pose.

\textbf{Reference Foot Contact Reward:}
Regarding the reference foot contact reward in Table \ref{reward}, learning different gaits requires different reference contact patterns. For multi-skill training, we activate different gaits according to the relative distance $d$ between the goal and the robot as described in the target-following reward. Specifically, we use the trotting contact pattern as reference to calculate this reward if $|d|<x_1$, bounding contact pattern if $x_1\leq |d|<x_2$, and gallop contact pattern if $|d| \geq x_2$, where $x_1$ and $x_2$ are the gait switch criteria we will automatically discover in Section \ref{method2} via an optimization loop outside the motor learning progress. Besides, it should be noted that for single skill policies, trotting has a similar velocity range as pacing and is a more common and stable gait. Therefore, when training our multi-skill policy, trotting was activated by the related reward terms when the goal is close rather than pacing. Nevertheless, technically, pacing can also be included for training a new multi-skill policy if needed.

\section{Discovery of skill switching criteria}
\label{method2}
Here we propose to set up an optimization problem in the outer loop of the motor learning process to automatically discover the gait switch criteria from trotting to bounding and from bounding to galloping.
We use covariance matrix adaptation evolution strategy (CMA-ES) \cite{hansen2016cma}, which is a derivative-free evolution strategies inspired by biological evolution for optimization. We aim to find gait switch criteria $x_1 m, x_2 m$ to maximize the sum of target following reward $r_g$ over each episode. The optimization problem is formulated as:
\begin{align*}
\arg\min_{x_1, x_2} & \sum_{i=1}^N -r_g(x_1,x_2) \\
\text{s.t.} \quad & 0 \leq x_1 \leq 15, \\
                 & 0 \leq x_2 \leq 15, \\
                 & x_1 < x_2.
\end{align*}
where $x_1,x_2$ are the decision variables representing the relative distance between robot and goal in the horizontal plane to activate the switch between trotting and bounding, bounding and galloping gaits, respectively. $N$ is the number of time steps in an episode.
\section{Experimental results}
\label{results}

This section first introduces experimental setup and then presents the optimization process of finding the proper gait switch criteria. 
We then demonstrate that the proposed multi-skill policy can achieve versatile locomotion skills by giving normalized goal distance input. 
Moreover, we show that the robot can recover from failures at any time during multi-skill locomotion. 
Furthermore, we showcase our proposed approach outperforms the baseline, i.e., manual switch between single skills, in terms of stability, smoothness and robustness.

\subsection{Experimental setup}

\textbf{Multi-skill training setup.}
We sample 5000 steps from the composite multi-skill policy for each training epoch, i.e., 20 episodes without early termination; Each episode lasts 10 seconds; Batch size is 128; Replay buffer size is 1e6; Learning rate is 3e-4; Weight decay is 1e-6; Soft target update is 0.001; Discount factor is 0.995 for fall recovery and 0.955 for locomotion gaits.

\textbf{Skill switch criteria optimization setup.}
We set initial gait switch criteria as $x_1 = 2.0 m$ and $x_2 =5.0 m$ to warmstart the optimization, population size as 50, and $\sigma = 1.0m$. The CMA-ES optimization runs for an iteration for every 20 iterations of inner loop deep reinforcement learning.

\begin{figure}[h]
  \centering
  \includegraphics[height=1.8in]{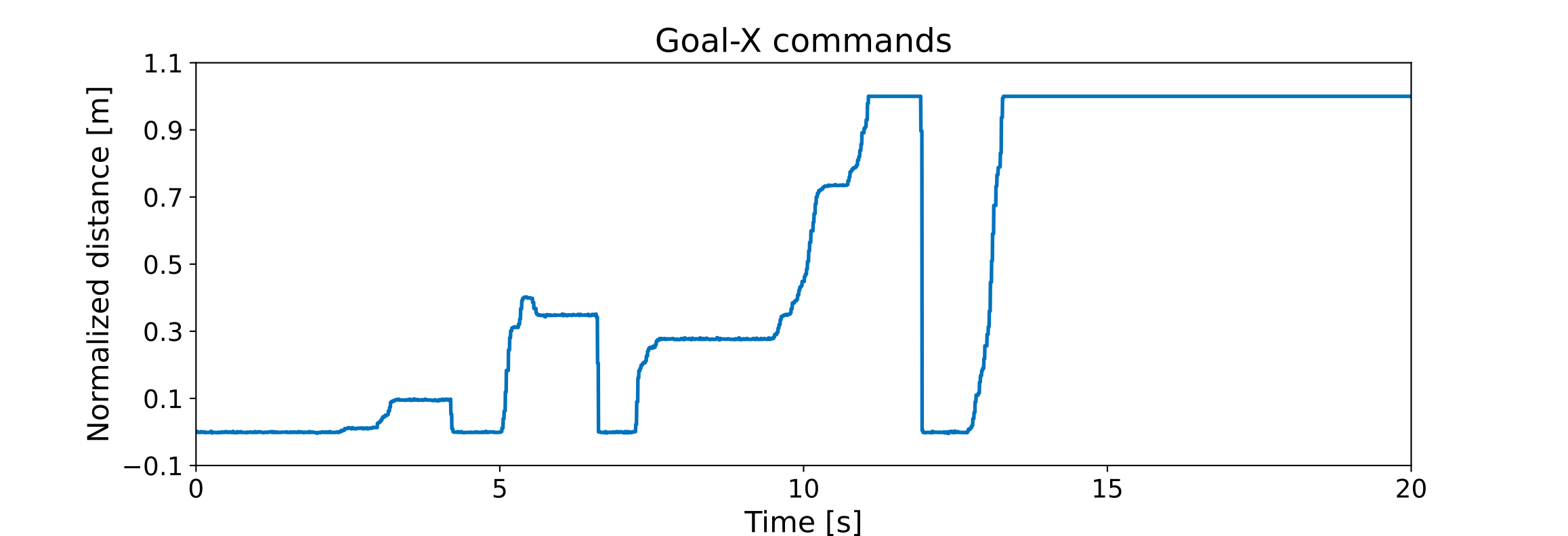}
  \caption{Normalized relative goal command in robot heading frame is given to encourage fall recovery, trotting, bounding, galloping and their transitions.}
  \label{fig:goal}
\end{figure}

\textbf{Goal trajectories setup.}
We give normalized relative goal distance in the x- and y-axis in the robot heading frame via a joystick in real-world tests to encourage the emergence of multiple dynamic skills and their transitions. An example goal trajectory is shown in Fig. \ref{fig:goal}.

\textbf{Velocity estimation.}
For the deployment of multi-skill policies learned in simulation on real robots, sensing errors and uncertainties usually cause the discrepancies between simulation and the real world. Unlike other state observations we selected for learning, base linear velocity cannot be obtained directly and needs to be estimated via leg kinematics or visual odometry, where the estimation results perform bad during foot slipping or highly dynamic motions \cite{ji2022concurrent}.
To this end, similar to \cite{ji2022concurrent}, we train a separate velocity estimator, to obtain the estimation of unavailable or unreliable states given the sensory information of more reliable states.
The input to the state estimator is 66-dimensional including gravity vector from roll and pitch measurements from IMU, two-step history of gravity vector, base angular velocity from IMU, two-step history of base angular velocity, joint positions, two-step history of joint positions, and joint velocity from motor encoders.
The output is 3-dimensional estimated base linear velocity.
The estimator network is composed of two hidden layers each has 256 neurons and uses a ReLU activation function.
After we obtain the locomotion policies in simulation, we collect 215000 pairs of input-output for training the estimator network via supervised learning.
We use mean squared error loss (ground truth velocity versus velocity estimation generated by the neural network) for training, where learning rate is 0.001, weight decay is 0.0005, and batch size is 1024.

\subsection{Optimized skill switch criteria}
\begin{figure} [h]
    \centering
    \includegraphics[height=2.2in]{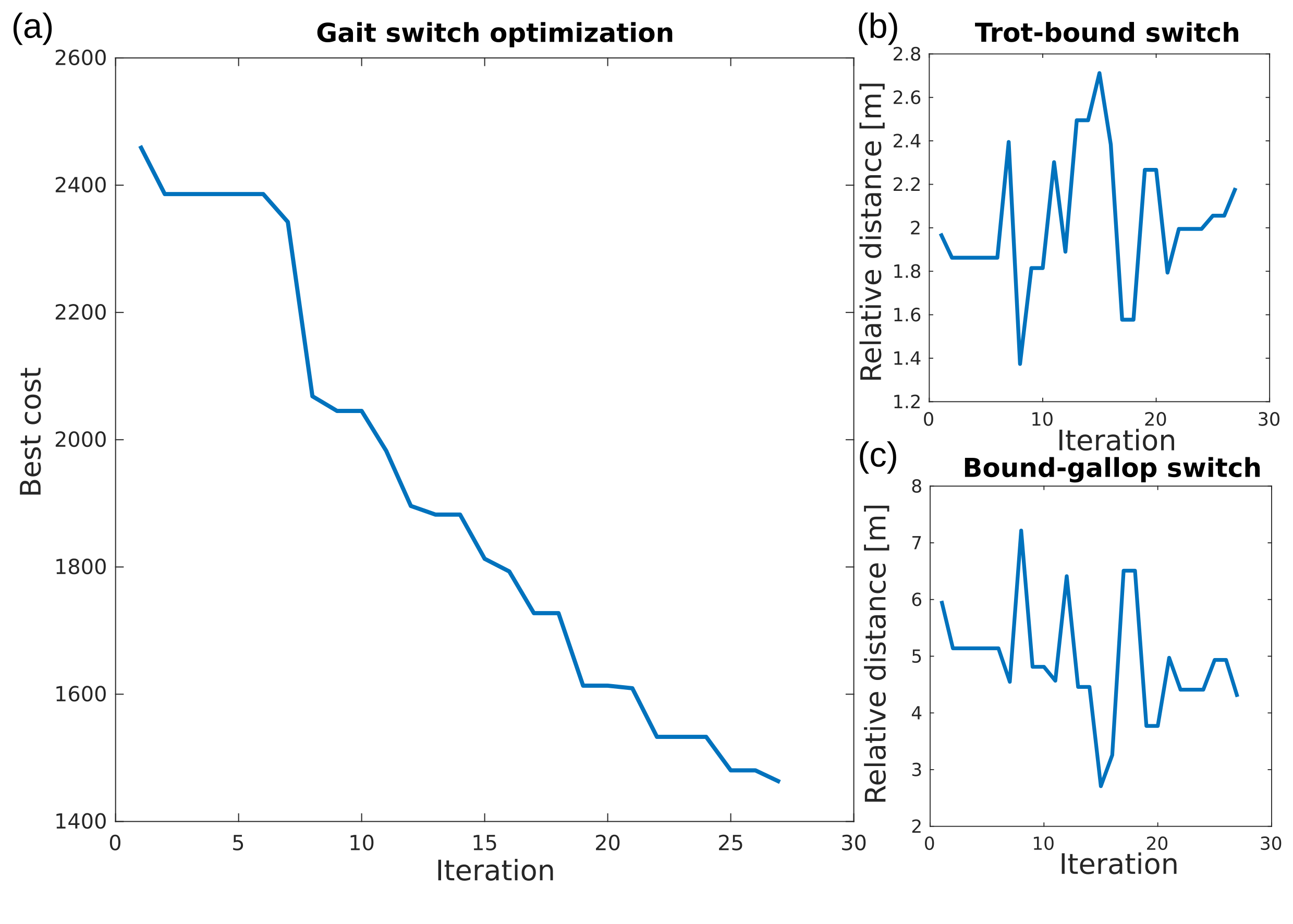}
    \caption{Results of CMA-ES optimization for gait switch criteria in learning multi-skill locomotion. (a) Best cost during CMA-ES optimization. (b) Optimized relative distance for switching from trotting to bounding. (c) Optimized relative distance for switching from bounding to galloping.}
    \label{fig:cma}
\end{figure}

From Fig. \ref{fig:cma}, we can see that the best cost value keeps decreasing and reaches a local minimum value after 27 iterations of optimization, where the corresponding trot-bound and bound-gallop switch criteria are $x_1 = 2.2 m$ and $x_2 = 4.3 m$, respectively. It should be noted that the optimized gait switch criteria are not exactly where the gait transitions occur in practice since they are only incorporated in reward functions. Instead, the gait transitions during multi-skill locomotion is naturally learned via the optimized gait switch criteria.

\subsection{Multi-skill locomotion with continuous skill transitions}
\begin{figure}[h]
  \centering
  \includegraphics[height=3.5in]{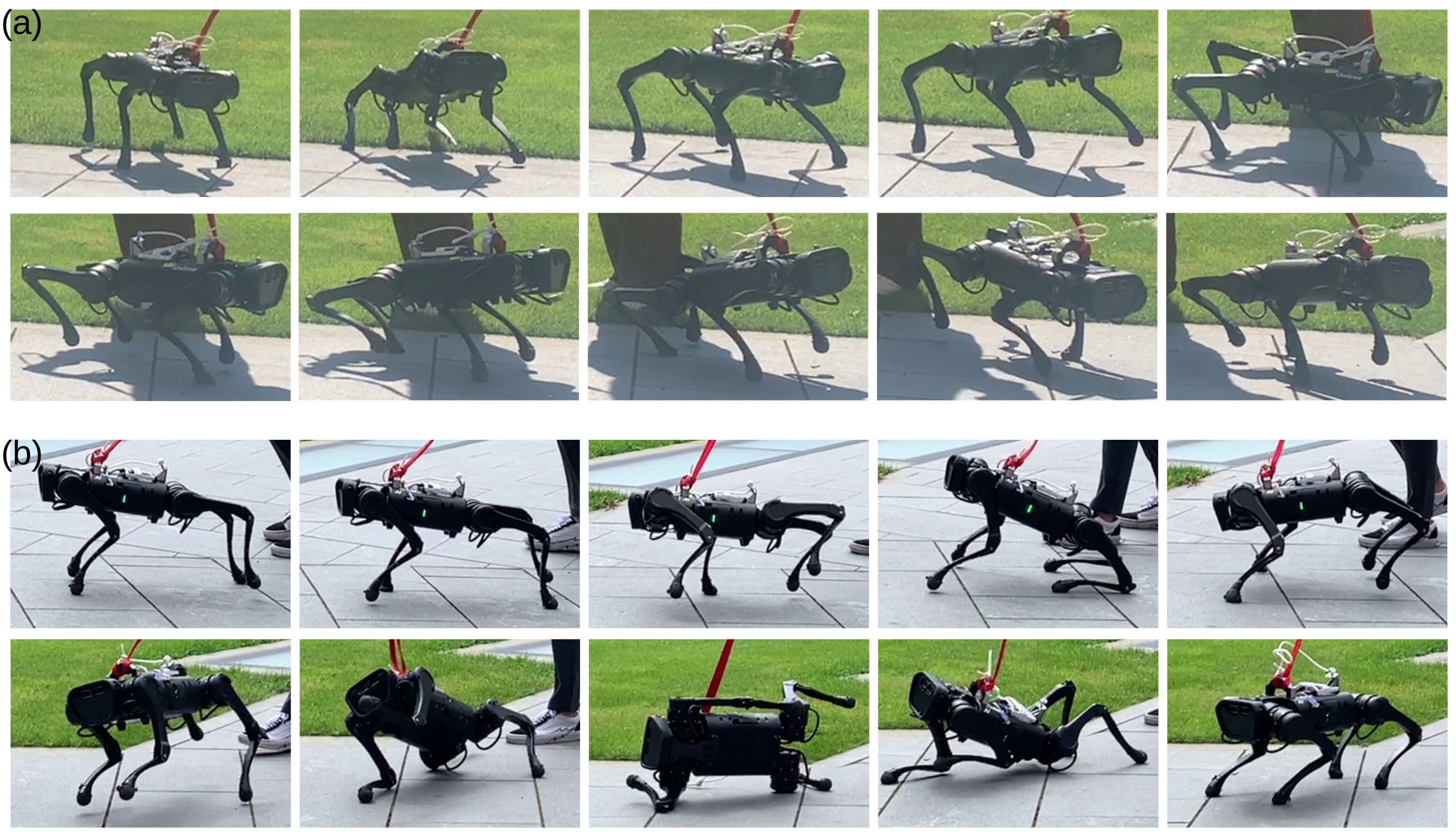}
  \caption{Comparison of multi-skill locomotion using our proposed approach and baseline. (a) Learned multi-skill locomotion by following goal trajectories in Fig. \ref{fig:goal}. (b) Baseline approach by manually switching between learned single skills. The robot failed after a discrete switch from bounding to galloping.}
  \label{fig:snapshots_goals}
\end{figure}

With the learned multi-skill locomotion policy, the robot is able to demonstrate trotting, bounding, galloping, and prompt fall recovery whenever necessary, as shown in Fig. \ref{fig:snapshots_goals}(a) and the accompanying video.
The corresponding goal commands given via joystick is shown in Fig. \ref{fig:goal}. Here we only report the experimental results in the real world. Refer to the accompanying video for more simulation and robustness tests.

\begin{figure}[h]
  \centering
  \includegraphics[height=4.0in]{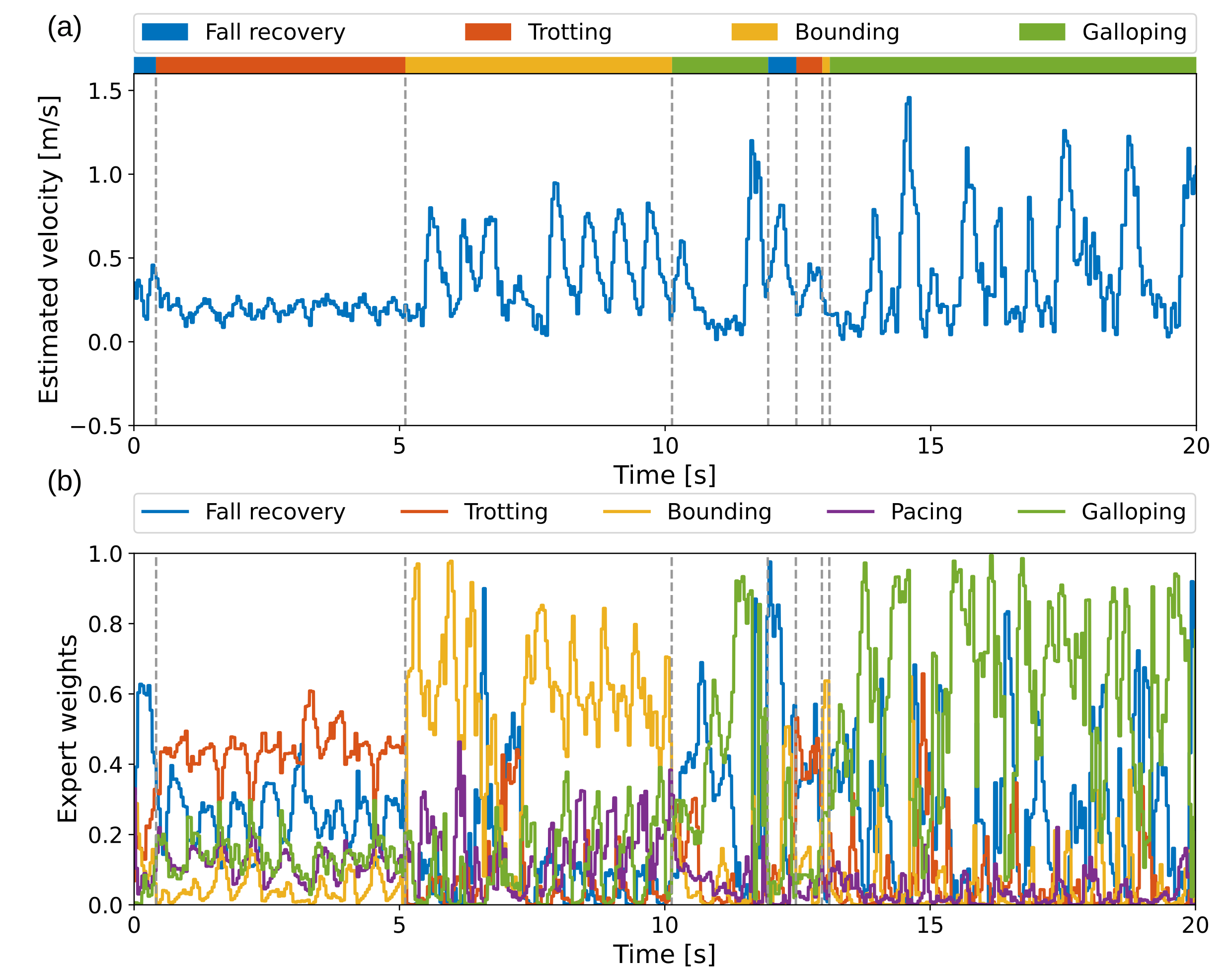}
  \caption{Versatile locomotion with continuous gait transitions. (a) Estimated horizontal speed of the robot during multi-skill locomotion, and (b) Expert weights showing each motion utilizes all the five experts with one related expert dominates.}
  \label{fig:goal1_speed_weights}
\end{figure}

\textbf{Estimated speed.}
We show the estimated speed in the horizontal plane of the robot heading frame for 20 seconds of the multi-skill locomotion, see Fig. \ref{fig:goal1_speed_weights}(a). The robot shows an increasing velocity for trotting, bounding and galloping skills.

\begin{figure}[h]
  \centering
  \includegraphics[height=2.5in]{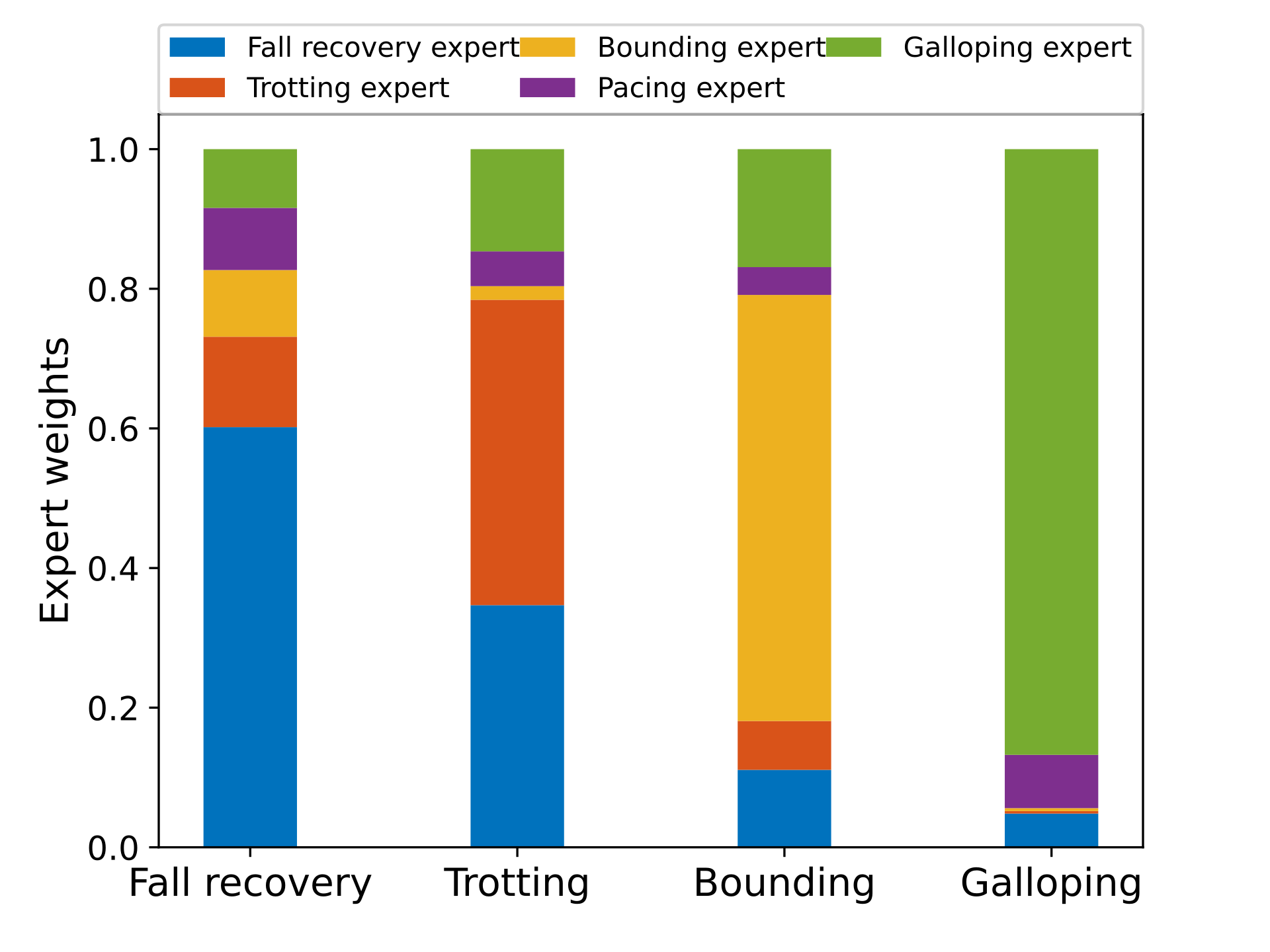}
  \caption{Composition of five skill primitives for fall recovery, trotting, bounding and galloping during multi-skill locomotion at 0.2s, 2.5s, 8.0s and 18.0s, respectively.}
  \label{fig:goal1_weight_bar}
\end{figure}

\textbf{Expert weights.}
Figure \ref{fig:goal1_speed_weights}(b) shows the weights for each single-skill policy generated by the gating network correspondingly.
The fall recovery expert dominates during the recovery motion. For the three locomotion gaits exhibited, each corresponding expert has the largest weight among all. However, compared to bounding and galloping gaits where the corresponding expert dominates, the trotting expert is co-acting more together with other experts during trotting in contributing to the synthesized policy. 
Furthermore, to have a detailed view of the influence of each expert in different quadruped locomotion skills, we visualized the weight for each expert of the four demonstrated skills at different time steps. From Fig. \ref{fig:goal1_weight_bar}, we clearly see the composition of single-skill policies for each motion, and each motion utilizes all the five single skills.

\begin{figure}[h]
    \centering
    \includegraphics[height=2.0in]{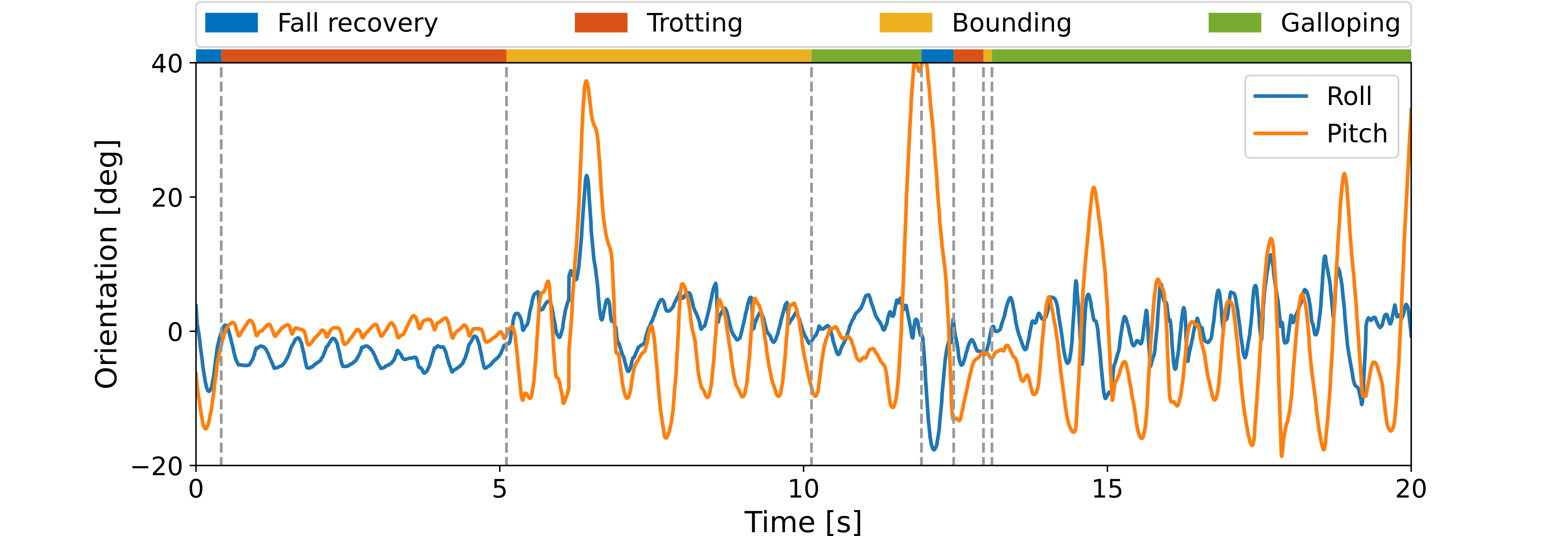}
    \caption{Roll and pitch angles during multi-skill locomotion.}
    \label{fig:orn}
\end{figure}

\textbf{Euler angles.}
The corresponding roll and pitch angles are shown in Fig. \ref{fig:orn}. When the robot encountered falls, the roll and pitch angles could become large at first and return to normal range during fall recovery. In other case, these two euler angles have clear cyclic pattern. Moreover, from trotting to bounding to galloping, the magnitude of the euler angles becomes larger, indicating the motion becomes more dynamic.

\subsection{Ablation studies}
We compare our proposed multi-skill learning and optimization approach with baseline approach, i.e., manual switch between different skill primitives. 
For the single skills, trotting and bounding were learned with a fixed desired velocity, while galloping was learned by maximizing velocity.
After multi-skill learning with the parameters of each expert network fixed, trotting motion is synthesized by the gating network at a lower speed range, and galloping motion is synthesized in a more dynamically feasible pattern with a higher success rate of real-world deployment. See the snapshots in Fig. \ref{fig:snapshots_goals} and the accompanying video for more details.

\begin{figure}[h]
  \centering
  \includegraphics[height=4.0in]{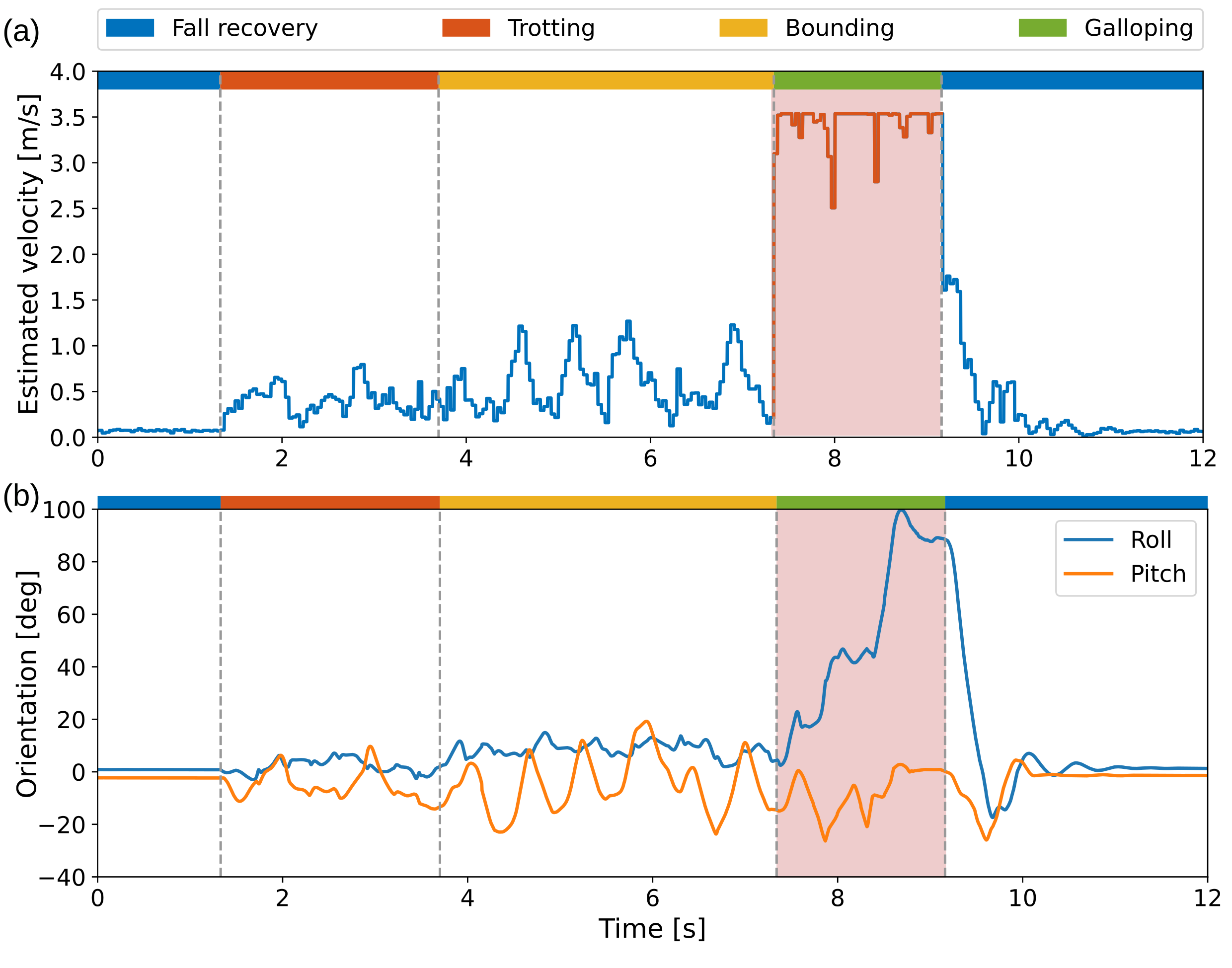}
  \caption{Performance of baseline approach by manually switching from fall recovery to trotting to bounding. Robot failed to switch to galloping from bounding (red shaded areas), but was then able to perform a successful recovery from failure to standing still. (a) Estimated horizontal speed. (b) Roll and pitch angles.}
  \label{fig:baseline}
\end{figure}

For the baseline approach, the robot failed when manually switching from bounding to galloping, sometimes causing automatic shutdown due to the power protection for Unitree robots. In the case when it failed without triggering the power protection, we can manually activate fall recovery skill, and the robot can then recover from failure to standing still. However, to resume locomotion, it requires another discrete switch from standing to trotting, bringing further instability issues. In contrast, our multi-skill policy can directly transition from failure to trotting without the intermediate phase in a dynamic fashion.
When discretely switching from trotting to bounding at an improper gait phase, the knee joints of the rear legs may become very close to the ground or the front legs lift very high in the following several time steps. 
As shown in Fig. \ref{fig:baseline}, manual switch caused dynamically instability, such as abrupt changes in estimated velocity. While our approach enables a smoother and continuous gait transitions in the real-world deployment.

\clearpage
\newpage
\section{Conclusions}
\label{conclusion}

This research developed a hierarchical learning and optimization framework to achieve multi-skill locomotion with optimized gait switch criteria without the need for any reference trajectories or expert demonstrations. The robot demonstrates continuous gait transitions among trotting, bounding and galloping skills as locomotion speed increases. Our learned multi-skill policy can also incorporate the fall recovery skill which enables the robot to recover promptly and resume locomotion whenever it becomes unstable or falls during different gaits. Thus, the robot requires less human intervention to operate autonomously in a remote working space, enabling versatile applications. Compared with the existing end-to-end learning framework using a single policy, our hierarchical framework is bio-inspired and more efficient in fine-tuning for various tasks since it does not require learning from scratch to adapt to new tasks. 
Moreover, by optimizing gait switch criteria as motor learning progresses, we avoid manually specifying the criteria with biased human knowledge distillation. The formulation can be easily adapted and generalized to different tasks or scenarios by customizing the cost function.

One limitation of our approach is that we found sim-to-real discrepancies still exist in galloping motion. In simulation, the galloping is more stable without any failures. In the real world, we cannot ensure 100 \% success rate of galloping for very long periods.
This can be due to various reasons.
Compared to other locomotion skills, galloping is inherently a very unstable locomotion skill, since only one foot is in contact with the ground at one time. Slight sim-to-real discrepancies can cause huge difference and even failures, such as deformable foot pads on Unitree A1 robot, ground friction, velocity estimation for out-of-the-distribution motion.
Another reason is the goal command we give in real-world tests is different from the simulation. For training in simulation, we give goal position directly in the world frame, while in the real world, due to the lack of body position feedback, we give normalized relative goal distance in robot heading frame via joystick, which is not possible to reproduce the same goal commands as in simulation via joystick.

For future work, we plan to further resolve the sim-to-real gap in galloping motion. We would also like to investigate the scalability of the proposed framework to more skills and more complex tasks as it would need more reward engineering. 
Furthermore, since our proposed approach can generate multi-skill locomotion data without any reference, one interesting application of our proposed framework would be preparing datasets for the training and fine-tuning of the generalist polices for legged robots, such as diffusion models or OpenVLA \cite{kim2024openvla}.

\subsubsection*{Acknowledgments}
This work was supported by the UKRI Future Leaders Fellowship [MR/V025333/1] (RoboHike). Fernando Acero is supported by the UKRI CDT in Foundational Artificial Intelligence (EP/S021566/1). 
\clearpage
\newpage
\bibliographystyle{apalike}
\bibliography{jfrExampleRefs}

\end{document}